\documentclass[10pt,twocolumn,letterpaper]{article}

\usepackage{ijcb}
\usepackage{times}
\usepackage{epsfig}
\usepackage{graphicx}
\usepackage{amsmath}
\usepackage{amssymb}
\usepackage{makecell}
\usepackage{url}
\usepackage[affil-it]{authblk} 

\ijcbfinalcopy 

\ifijcbfinal\fi
\begin{document}

\title{DFGC 2021: A DeepFake Game Competition}

\author{\normalsize
Organizers \\
Bo Peng\textsuperscript{1}, Hongxing Fan\textsuperscript{1}, Wei Wang\textsuperscript{1}, Jing Dong\textsuperscript{1}, Yuezun Li\textsuperscript{2}, Siwei Lyu\textsuperscript{3}, Qi Li\textsuperscript{1}, Zhenan Sun\textsuperscript{1} }
\affil{
\textsuperscript{1}CRIPAC, NLPR, Institute of Automation, Chinese Academy of Sciences \\
\textsuperscript{2}Department of Computer Science and Technology, Ocean University of China \\
\textsuperscript{3}University at Buffalo, State University of New York, NY14260, USA 
}
\author{\normalsize
Competition Winners \\
Han Chen\textsuperscript{4}, Baoying Chen\textsuperscript{4}, Yanjie Hu\textsuperscript{4}, Shenghai Luo\textsuperscript{4}, Junrui Huang\textsuperscript{5}, Yutong Yao\textsuperscript{5}, Boyuan Liu\textsuperscript{5},
Hefei Ling\textsuperscript{5}, \\
Guosheng Zhang\textsuperscript{6}, Zhiliang Xu\textsuperscript{5}, Changtao Miao\textsuperscript{7}, Changlei Lu\textsuperscript{7},
Shan He\textsuperscript{8}, Xiaoyan Wu\textsuperscript{8}, Wanyi Zhuang\textsuperscript{7}
}
\affil{
\textsuperscript{4}Shenzhen Key Laboratory of Media Information Content Security, Shenzhen University, China 
\\
\textsuperscript{5}Huazhong University of Science and Technology,	1037 Luoyu Road, Wuhan, China
\\
\textsuperscript{6}Guangdong University of Technology,	No.100 Waihuanxi Road, Guangzhou, China
\\
\textsuperscript{7}University of Science and Technology of China,Hefei, China  \quad \textsuperscript{8}iFlytek Research, iFlytek Co., Ltd.
}

\maketitle
\thispagestyle{empty}

\begin{abstract}
   This paper presents a summary of the DFGC 2021 competition\footnote{\url{https://competitions.codalab.org/competitions/29583}}. DeepFake technology is developing fast, and realistic face-swaps are increasingly deceiving and hard to detect. At the same time, DeepFake detection methods are also improving. There is a two-party game between DeepFake creators and detectors. This competition  provides a common platform for benchmarking the adversarial game between current state-of-the-art DeepFake creation and detection methods. In this paper, we present the organization, results and top solutions of this competition and also share our insights obtained during this event. We also release the DFGC-21 testing dataset collected from our participants to further benefit the research community\footnote{\url{https://github.com/yuezunli/celeb-deepfakeforensics}}.
\end{abstract}

\section{Introduction}
DeepFake creation and detection are on-going adversarial games. With DeepFake results becoming more realistic and hard to be distinguished by human eyes, malicious DeepFake creators may also target on decieving automatic DeepFake detection models, bringing new challenges for current detection methods.

There has been some related DeepFake detection competitions, but to our knowledge, the creation side and the detection side have not been organized together in a multi-phase game environment, where each side is encouraged to defeat the other side as much as possible. Some related DeepFake competitions or benchmarks are compared with the DFGC in Table \ref{tabCompetitions}, including FaceForensics++ Benchmark \cite{FF++}, DeepFake Detection Challenge (DFDC) \cite{DFDC}, and DeeperForensics Challenge \cite{jiang2021dfc20}. 

The FaceForensics++ Benchmark \cite{FF++} is the first publicly available benchmark for evaluating DeepFake detection methods. Participants run their detection methods on its publicly available test set consisting of 1000 images and submit their prediction results to the evaluation server. This kind of evaluation is relatively easy to implement, but has the risk of human labeling as the test data is public.  

DFDC \cite{DFDC} is currently the most large-scale DeepFake competition, which is organized by Facebook with 1 million USD awards. It provides a large training dataset with 128,154 consented videos. It has a two-stage evaluation mechanism, where the Test1 data is used for the first stage public leaderboard, acting as a development dataset, and the Test2 data is used for the final ranking, with additional organic data from the web.

DeeperForensics Challenge \cite{jiang2021dfc20} is also a two-stage  competition similar to DFDC. It encourages participants to use the DeeperForensics dataset \cite{jiang2020deeperforensics1} for training their models, but external public datasets are also allowed for training.
\begin{table*} [thb] 
\centering
\caption{Recent DeepFake detection competitions and benchmarks.}  \label{tabCompetitions}
\renewcommand{\arraystretch}{1.0}
\scalebox{0.8}{%
\begin{tabular}{|c|c|c|c|c|c|c|c|}
  \hline
 Events   & Year   & Submission     & Test Data &  Fake Methods  & \makecell{Evaluation \\Metric}     & Train Data     & \makecell{Adversarial  \\Game} \\
  \hline
  \hline
  \makecell{FaceForensics++ \\Benchmark \cite{FF++} }     	 
  &2019  & Result   & 1000 Images   & \makecell{4 Methods, \\post processing}    &Acc    & Unrestricted  & No \\
  \hline
  DFDC  \cite{DFDC}
  &2020  & Code   & \makecell{Test1: 4000 videos \\Test2: 10000 videos}   & \makecell{8 Methods, \\ post processing, \\ additional organic data}    &Logloss    & \makecell{DFDC train set, \\External public datasets}  & No \\
  \hline
  \makecell{DeeperForensics \\Challenge \cite{jiang2021dfc20}}
  &2020  & Code   & \makecell{Test1: 1000 videos \\Test2: 3000 videos}   &\makecell{Unkown wild data, \\ post processing}   &Logloss    & \makecell{DeeperForensics dataset, \\External public datasets}  & No \\
  \hline
  DFGC 
  &2021  & Code   & \makecell{1000 real images \\N $\times$ 1000 fake images}   & Created by participants    &AUC    & \makecell{Restricted to CelebDF-v2 \\ train set \& its derivation}  & Yes \\
  \hline
\end{tabular} 
}
\end{table*}
\begin{figure*}[thb]
\centering
\centerline{\includegraphics[width=18cm]{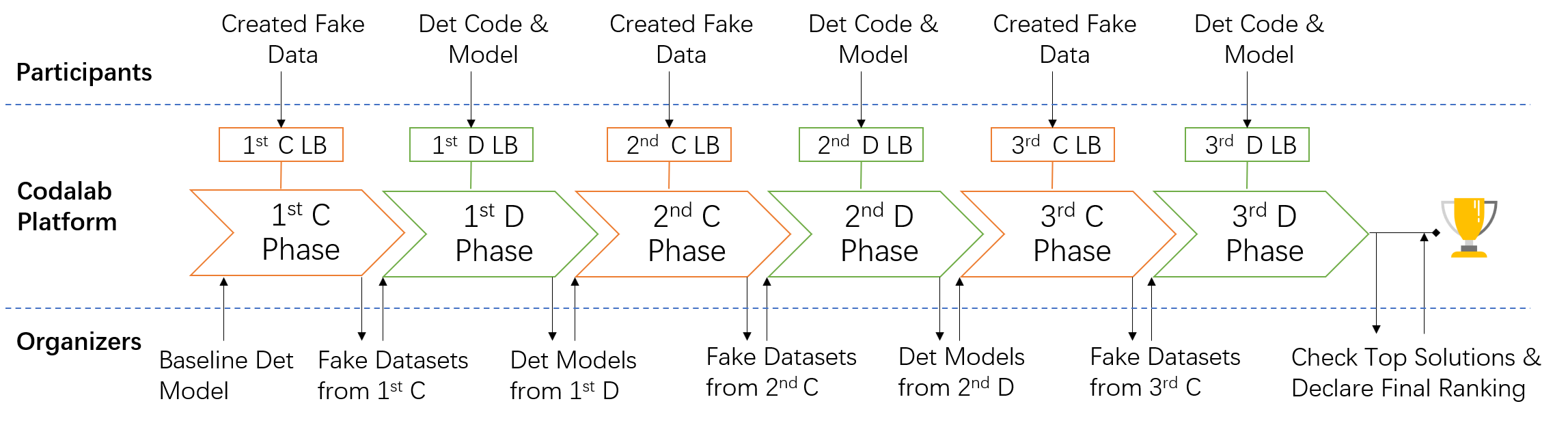}}
\caption{The competition consists of 6 alternating DeepFake creation and detection phases that are adversarially evaluated against their previous ones, where ``C" is for creation, ``D" is for detection, and ``LB" stands for leaderboard. }
\label{fig_workflow}
\end{figure*}


There were also some competitions that have similar adversarial form with ours. Our competition is partly inspired by two recent competitions held  in China \cite{AliCompetition, GeekPawn}, which both evaluate the fake creation and detection sides against each other. Differently, the Alibaba Security challenge \cite{AliCompetition} targets on the localization of certificate image forgery, and the GeekPawn DeepFake challenge  \cite{GeekPawn} does not have quantitative metrics for evaluating DeepFake image quality. More over, our competition has 6 alternating creation and detection phases, forming a multi-phase dynamic game. 

In summary, current DeepFake detection methods are usually evaluated on several established datasets, e.g. \cite{FF++, CelebDF, jiang2020deeperforensics1}, but they are not fully tested under intentional attacks from the DeepFake creation side. For the other side, novel DeepFake creation methods are developing fast, but they only focus on visual quality while their undetectability to detection methods are not tested. Hence, we propose the DFGC to provide a common platform for benchmarking the adversarial game between current state-of-the-art DeepFake creation and detection methods. 

\section{The DFGC}
\subsection{Overall Design}
From the DeepFake detection perspective, unknown fake methods and adversarial measures may increase detection difficulty. 
From the DeepFake creation perspective, participants need to create face-swap results that can most deceive unknown detection models and be of high quality in the same time. 
A multi-phase game gives participants the chance to test their strategies and inferring the counter-party's strategies.
These requires our competition to be very different from existing competitions or benchmarks in competition form and protocols.
 
The overall workflow is shown in Fig. \ref{fig_workflow}. The competition is composed of 6 interleaved phases for DeepFake creation and DeepFake detection. Each phase lasts for one week. During each creation phase (C-phase), participants can submit their created DeepFake datasets following some protocols in \ref{subsec_dataset}, and then obtain the creation evaluation results. During each detection phase (D-phase), participants can submit their detection codes (and models), and then obtain the detection evaluation results. Each phase has its own real-time leaderboard (LB), which shows only one score (typically the best performing score) from each team.

To form an adversarial game, the detection methods submitted in a D-phase are evaluated against all created DeepFake datasets that are submitted to its previous C-phase leaderboard. The creation datasets submitted in a C-phase are evaluated against all DeepFake detection methods submitted to its previous D-phase leaderboard. Note the created datasets are also evaluated by some image quality metrics, which will be detailed in \ref{subsec_metric}. The organizers update the evaluation codes and datasets/models when switching phases as shown in Fig. \ref{fig_workflow} to induce the multi-phase dynamic game, where each party is evaluated against an updated and more challenging counter-party.

The competition is hosted on the Codalab platform, which is a popular open-source platform for hosting academic competitions. The organizers setup the evaluation codes, reference datasets, and a backend GPU worker, and the Codalab platform automatically handles the processing of registration, submission and leaderboard etc. The organizers also update the evaluation codes and reference datasets/models at the switching of different phases as mentioned above.

\subsection{Datasets} \label{subsec_dataset}
For a controllable evaluation and a common basis for model training, we require both creation and detection tracks to base on a constrained dataset, for which we choose the Celeb-DF v2 dataset \cite{CelebDF}. This dataset is composed of 590 real videos and 5,639 high-quality fake videos that are created by a modified AutoEncoder based DeepFake method. The Celeb-DF dataset is split into a test set of 518 videos and a train set composed of the rest videos. Since current state-of-the-art DeepFake detection methods are mostly frame-based or integration of frame-based methods \cite{DFDC, jiang2021dfc20}, and also considering evaluation efficiency, our evaluations are conducted on video frames instead of video clips.

For the creation track, we specify 1000 face-swap images to be created and submitted by participants. These images are specified from the Celeb-DF test set fake videos. They are named as ``idT\_idS\_vidIdx\_frameIdx.png", which means the \textit{frameIdx}-th frame of a target person's video \textit{idT\_vidIdx.mp4} needs to be swapped to the source person \textit{idS}'s face. The created face-swap image has the same image size and background as the target image ``idT\_vidIdx\_frameIdx.png" while the facial ID is swapped to \textit{idS}. Following this protocol, a dataset can be directly extracted from the test set of Celeb-DF, and submitted as a baseline dataset for evaluation. Creation track participants are encouraged to submit newly-created face-swap results, and they can also add post-processing and adversarial noise to make their datasets more challenging to the detection track.

For the detection track, participants are restricted to train their model only on the Celeb-DF train set and no external dataset is allowed. It is also permitted to augment the train set with re-created or post-processed data as long as they are created using the data resources of Celeb-DF train set. This rule is to maintain a common data basis for model training. Submitted detection methods (models) are evaluated on a 1000 real image set from Celeb-DF test set and all $N \times 1000$ testing fake images submitted to the previous C-phase LB, where $N$ is the number of submissions.

\subsection{Evaluation Metrics} \small  \label{subsec_metric}
The evaluation of detection submissions is straightforward. We use the mean AUROC value to measure the overall performance to discriminate real and fake samples. The detection score $S_D$ is calculated as:
\begin{equation} \label{eqn_SD}
    S_D = \sum_{i=1}^{N_C} \text{AUROC}_i / N_C
\end{equation}
where $N_C$ is the number of submissions on the previous C-phase LB, and $\text{AUROC}_i$ is the AUROC value calculated on the $i$th submitted fake dataset versus the real dataset, both of which have 1000 images.

The evaluation for the creation phase has more aspects to be considered. In the context of our competition, a good DeepFake dataset should be a valid face-swap result, of high quality, and also deceiving to detection methods. To quantify these characteristics, we designed the following creation score, that is the sum of an ID-similarity score, an image-level similarity SSIM score, a noise score, and an anti-detection score:
\begin{equation} \small \label{eqn_SC}
\begin{split}
    & S_C = 
    \sum_{i=1}^{N}\text{SSIM}(I^{fak}_i, I^{tar}_i)/N  +\sum_{i=1}^{N} \text{Noise}(I^{fak}_i)/N \\
    &  + \sum_{i=1}^{N}\cos{(f_{id}^i, f_{id}^{src_i})}/N 
    + 2 \times \sum_{i=1}^{N_D} (1-\text{AUROC}_i)/N_D
\end{split}
\end{equation}
where $I^{fak}_i, I^{tar}_i$ are respectively a fake image and its corresponding target image, $N$ is the number of submitted fake images (i.e. 1000), $\cos(,)$ is the cosine similarity, $f_{id}^i$ is the ID feature of the image $I^{fak}_i$ extracted by a face recognition network \cite{vggface2}, $f_{id}^{src_i}$ is the average ID feature of the corresponding face swap source person, $\text{SSIM}(,)$ calculates image similarity, $\text{Noise}()$ penalizes noisy images using a noise level estimation method \cite{noiseLevel}\footnote{The noise score is added at the beginning of the second C-phase.}, $\text{AUROC}_i$ is the AUROC value of running the $i$th detection method on this fake dataset versus the real set, and there are $N_D$ of these detection methods submitted to the previous D-phase LB.

The final ranking for the detection track is determined by the final D-phase LB. Whereas the final ranking for the creation track is determined by rescoring against the final D-phase detection methods. This setup is to simulate real-world situations where creation typically comes first and detection comes later. Moreover, the organizers check top detection solutions for reproducibility and validity after the competition ends to ensure their abidance to the rules.

\subsection{Statistics and Results}
In total, more than 180 individuals applied to the competition in its duration of six weeks, and the statistics of each phase are listed in Table \ref{tabStatistics}\footnote{Listed C1 scores have been rescored by adding the noise score.} . 
As can be seen the degree of participation increased over time, indicated by the increasing numbers of submissions in total and on the LB. Each participant can make up to 10 submissions per day to fully test their ideas. For this competition, the detection track attracted more attention than the creation track, probably because creating high-quality DeepFake dataset is more time-consuming and complicated. 
\begin{table} [thb] 
\centering
\caption{Statistics of each phase. Note each team can only submit up to one submission result to be shown on LB.}  \label{tabStatistics}
\renewcommand{\arraystretch}{1.0}
\scalebox{0.75}{%
\begin{tabular}{|c|c|c|c|c|c|c|}
  \hline
    &C1    & D1   & C2     & D2      & C3    & D3 \\
  \hline
\#submission total  &19  &208  &100   &361   &397     &1050  \\
\hline
\#submission on LB  &4  &15 &10 &27 &21 &28 \\
\hline
\makecell{best score\\ on LB}  &1.56
   &0.98   &2.58   &0.85   &\makecell{2.83 \\ 2.42 (final)}   &0.94 \\
   \hline
\makecell{median score \\on LB}  &1.55 
   &0.85   &2.39   &0.68   &\makecell{2.37 \\ 2.15 (final)}   &0.60 \\
  \hline
\end{tabular} 
}
\end{table}

Although the C-phase best scores are not directly comparable given they are evaluated against different sets of detection methods, these scores monotonously increased. The two scores in the C3-phase are respectively the C3-LB score (evaluated against the 2nd D-phase models) and the final rescored score (evaluated against the 3rd D-phase models).
By observing submitted DeepFake datasets during the competition and querying the participants about their used creations methods, we obtain the following observations: \\
\begin{itemize}
  \item A large portion of submitted DeepFake data adds adversarial noise to deceive detection methods.
  \item At the final C-phase, nearly all submitted DeepFake datasets are created by some new methods different from the baseline Celeb-DF method.
  \item At the final C-phase,  a large portion of submitted datasets combine new creation methods and adversarial attack methods.
\end{itemize}
Since the baseline Celeb-DF fake images have relatively lower ID score, some new creation methods are used by participants to increase the ID score, and they may also be more effective in deceiving detection models. 

Observing the median scores in the three D-phases monotonously decreased, we can say that the submitted DeepFake datasets became more challenging during three C-phases for most detection track participants. However, D-phase best scores showed quite high performances, even when faced with various new DeepFake creation methods and adversarial attacks. Learning from top-3 detection solutions, they have the following characteristics worth-noting:
\begin{itemize}
  \item All top-3 detection methods use extensive data augmentation and re-creation to simulate various unknown DeepFake creation methods.
  \item Two out of three methods use adversarial examples to augment their training set to better detect adversarial attacked samples.
\end{itemize}
Note these augmentations and re-creations are conducted on the Celeb-DF train set, and no external dataset is used, which is in compliance with the competition rules.

\section{Top Solutions}
The final results and scores of each team can be found on the competition website\footnote{\url{https://competitions.codalab.org/competitions/29583#learn_the_details-final-results}}. In the following, we only introduce top-3 solutions in each track.
\subsection{Creation Track 1st}
\small{\emph{Members: Changtao Miao, Changlei Lu, Shan He, Xiaoyan Wu, Wanyi Zhuang}}

We choose Faceshifter [4] as the method to generate fake images. Based on the faceshifter model, we mainly adopt three strategies to improve the score. One is that we fine-tune the faceshifter model, and the other is to add adversarial noise to the forged image. Finally, in order to reduce the fraud area and noise level in the final forged image, we also designed a post-processing method.

\noindent \textbf{Pair-Specific Fine-tuning.}
The faceswap operation involves two characters, the source and the target. In order to get better faceswap effect, we train an exclusive model for each pair ($ID_{source}$, $ID_{target}$). Specifically, for each pair, we choose the videos corresponding to $ID_{source}$, and the videos corresponding to $ID_{target}$ from CelebDF dataset to build a sub-dataset, and then use this subdataset to fine-tune the faceshifter pre-trained model. The fake images generated by the fine-tuned model are clearer, and the id information is processed better, and the score in the game has been improved a lot, especially the anti-detection score.

\noindent \textbf{Adding Adversarial Noise.}
Based on the setting of the game, we also add adversarial noise to the fake images to improve the anti-detection ability. We first train a detector $D$ using the CelebDF dataset and the fake images generated by Faceshifter, where the backbone network is EfficientNet-B7. After that, we adopt the FGSM method \cite{goodfellow2014explaining}, and use the detector D to add noise to the required 1000 images.

\begin{figure}[h]
	\begin{center}
		\includegraphics[width=1.0\linewidth]{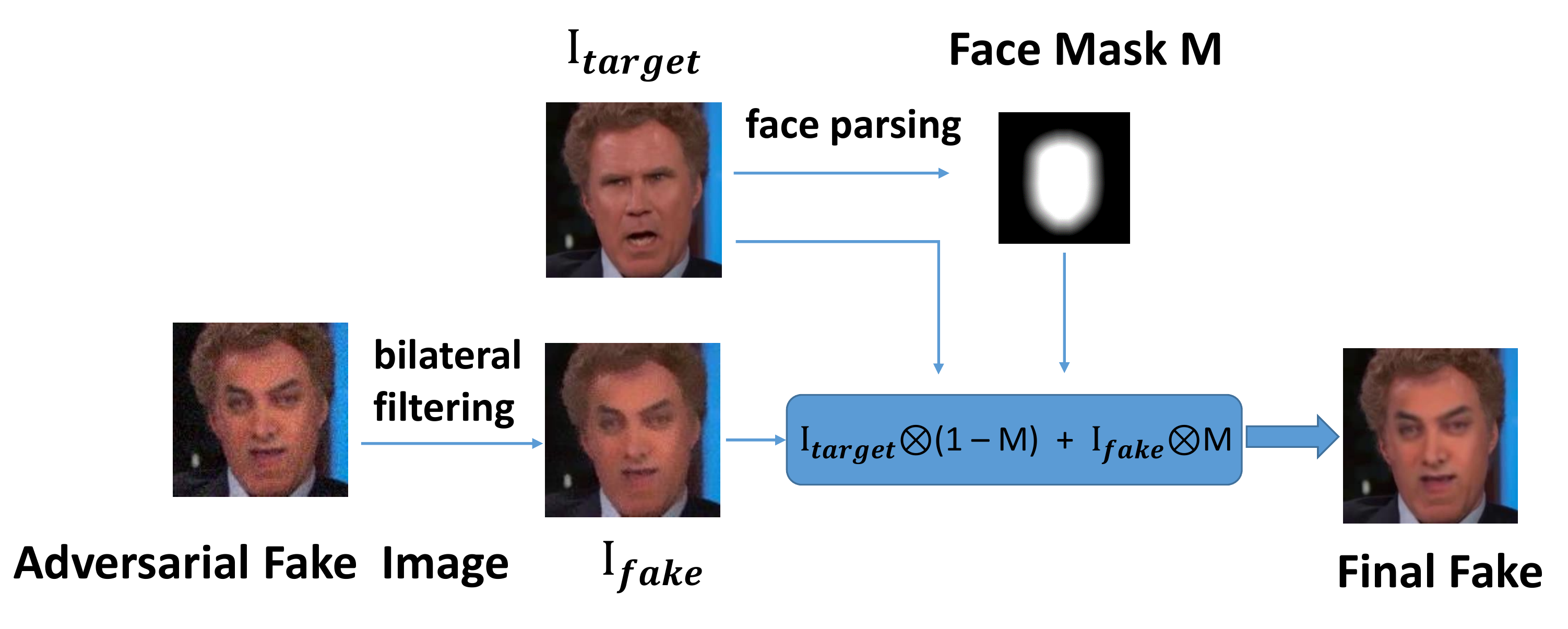}
		
	\end{center}
	\caption{Creation track 1st solution's post-processing of filtering and blending.}
	\label{fig:1stpp}
\end{figure}

\noindent \textbf{Post-Processing.}
In order to improve the noise score and SSIM score, we design this post-processing scheme. We first use a bilateral filter to filter the adversarial samples to improve the noise score in the game, and only fuse the face region in the fake image into the target image to get the final fake image. Fig.\ref{fig:1stpp} shows the corresponding post-processing flow. 

\subsection{Creation Track 2nd}
\small{\emph{Members: Junrui Huang, Yutong Yao, Boyuan Liu, Hefei Ling}}

\begin{figure}[h]
	\begin{center}
		\includegraphics[width=0.8\linewidth]{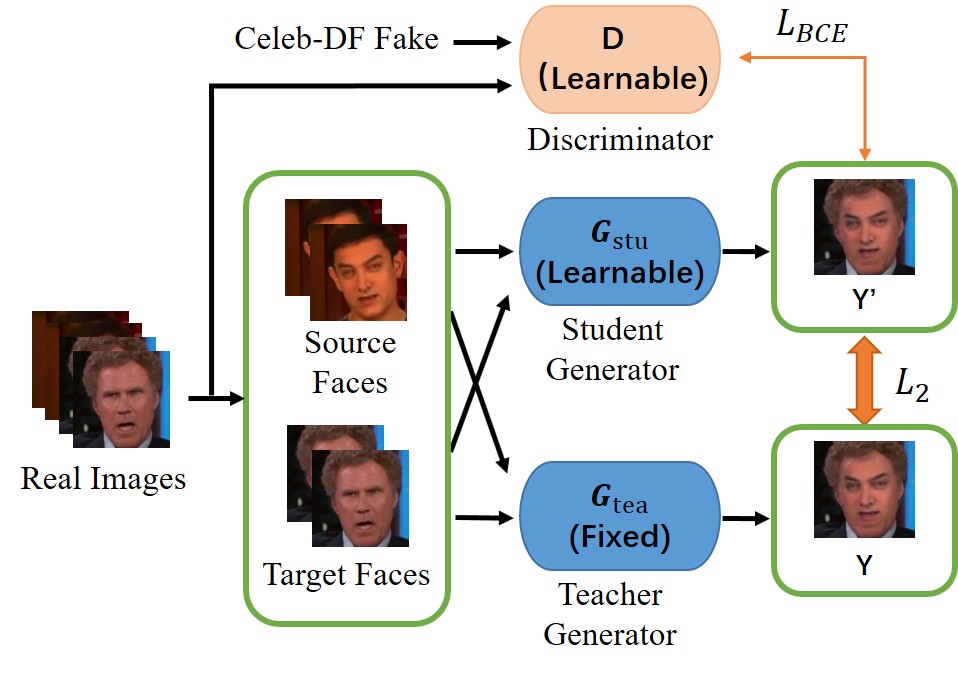}
		
	\end{center}
	\caption{Creation track 2nd solution's pipeline.}
	\label{fig:creationpipe}
\end{figure}

{\bf Main algorithm.} We use a pre-trained FaceShifter \cite{li2019faceshifter} model as the teacher model to guide the learning of a student FaceShifter model, and use an adversarial training method to improve the deceiving ability of the student model results. The whole framework is shown in Figure~\ref{fig:creationpipe}.  First, we apply $L_2$ loss to force the output of the student model to resemble the pre-trained teacher model. To better counter the detection track models, the student model is also adversarially trained against a discriminator that is trained on both generated samples and Celeb-DF fake samples. The total generator loss is:
\begin{equation}\small 
	L_{G} = L_{BCE}(D(Y'),0) +L_2(Y',Y)
	\label{loss_G}
\end{equation}
where the label 0 in the BCE loss represents real label, meaning student generation result $D(Y')$ needs to deceive the discriminator. $G_{tea}$  is the pre-trained teacher FaceShifter model and kept fixed during training.

For the adversarial training, the discriminator learns to classify samples $X_{real}$ and $X_{fake}$ in the Celeb-DF training dataset, and also student face swapping results $Y'$. The loss for the discriminator is defined in Eqn. \ref{loss_D}:
\begin{equation}\small 
	\begin{aligned}
		L_{D} = & L_{BCE}(D(Y'),1) +\\
		&L_{BCE}(D(X_{fake}),1) + L_{BCE}(D(X_{real}),0)
	\end{aligned}
	\label{loss_D}
\end{equation}
where the label 0 represents real and 1 represents fake.
The student model $G_{stu}$ learns to hide those forgery clues, thus resulting in better anti-detection ability than the teacher $G_{tea}$.

	\begin{figure}[h]
		\begin{center}
			\includegraphics[width=0.8\linewidth]{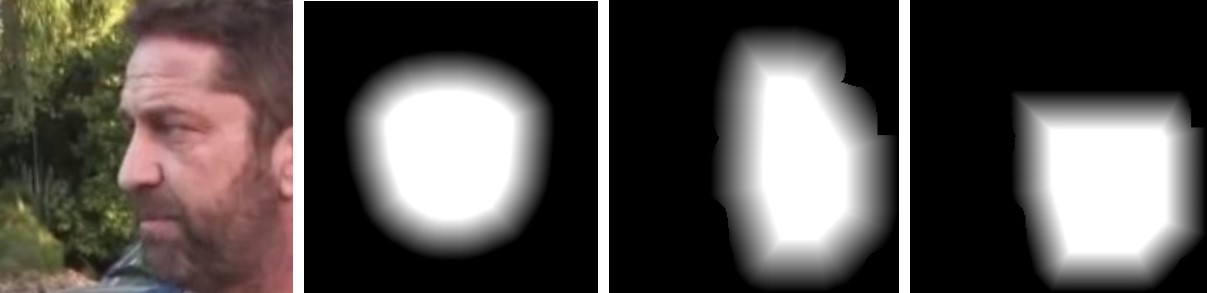}
			
		\end{center}
		\caption{Creation track 2nd solution's different face masks that are later manually picked for best blending quality.}
		\label{fig:mask-comparison}
	\end{figure}
	{\bf Quality tuning.}  To obtain better quality during face-swap creation, we pay more attention to source image selection and result blending. 
	We tend to select source faces which have similar poses as target faces to generate the face-swap results. This can minimize artifacts and face distortions caused by pose mismatch. 
	For result blending, we propose three types of face masks and select the best blending result for each specific sample. Considered masks are shown in Figure~\ref{fig:mask-comparison}. The first is a universal face mask that includes the central area of frontal face. The second is the mask that fits the edge of the face contour generated by face segmentation algorithm{\footnote{https://github.com/zllrunning/face-parsing.PyTorch}}. The third is the face-parsing mask without forehead region in order to avoid potential occlusions such as bangs and hats.

\subsection{Creation Track 3rd}
\small{\emph{Members: Guosheng Zhang, Zhiliang Xu}}

\begin{figure}[htb]
\centering
\includegraphics[width=1.0\linewidth]{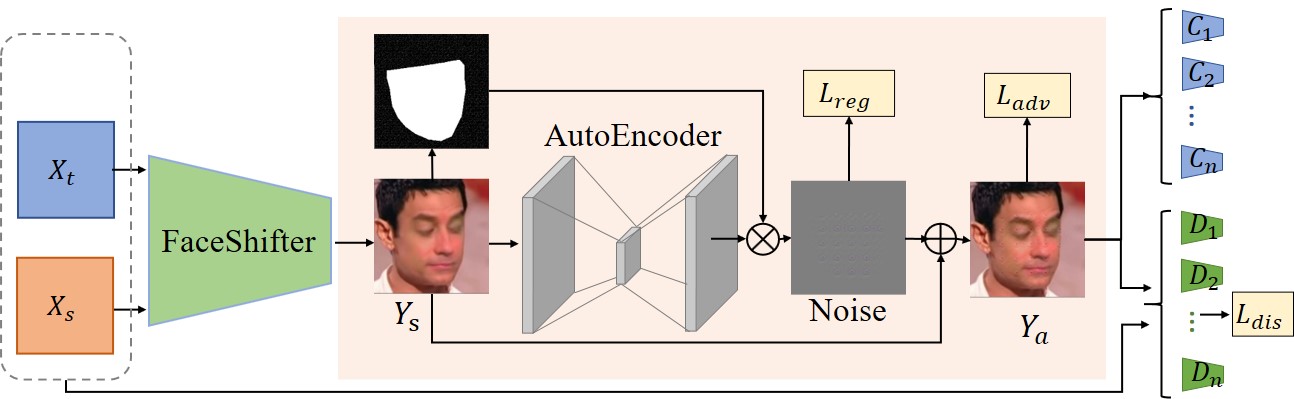}
\caption{Creation track 3rd solution.}
\label{zgs_fig}
\end{figure}

In this solution, we focus on creating face-swaps that can successfully attack deepfake detectors. To improve the attack ability of crafted examples, we propose a new adversarial training method against multiple discriminators with different network architectures. With adversarial training, the adversarial noises are added to face region to interfere with the deepfake detection models.
\\
{\bf Main algorithm.} The proposed adversarial training framework is illustrated Fig.~\ref{zgs_fig}. It  consists of a deepfake generator and an adversarial noise generator followed by multiple forensic classifiers $\{C_1, C_2,...,C_n\}$ and GAN discriminators $\{D_1, D_2,...,D_n\}$, where $n = 5$. We reproduce AEI-Net from the FaceShifter \cite{li2019faceshifter} method and use it as a baseline deepfake generator, and then we utilize an AutoEncoder network $E$ to generate adversarial noise in an adversarial training manner. To ensure that adversarial noise is capable of attacking robust 
forensic classifiers, we first train multiple forensic classifiers $\{C_1, C_2,...,C_n\}$ with different structures on newly generated Celeb-DF training data using FaceShifter \cite{li2019faceshifter}, FaceController \cite{xu2021facecontroller}, FaceSwap \cite{FaceSwap}, FirstOrderMotion \cite{siarohin2019first} and FSGAN \cite{nirkin2019fsgan}, apart from its original training data. 

Then we load the pretrained $C_i$ and initialize $D_i$ with the same parameters as $C_i$ to begin adversarial training. The difference between $C_i$ and $D_i$ is that $C_i$ is kept fixed while $D_i$ is adversarially updated during training.


As show in Fig.~\ref{zgs_fig}, We add adversarial noise to the original generated image $Y_{s}$ to generate the adversarial example $Y_{a}=E(Y_{s})\odot  mask + Y_{s}$, which aim to deceive $D_i$ and the pretrained $C_i$ . We formulate the adversarial loss as follows:
\begin{equation}\small  \label{eqn_C3rdLadv}
\min_{E}L_{adv}=\sum_{i=1}^{n} \text{log} D_i(Y_{a})+log C_i(Y_{a})
\end{equation}
Note we have omitted the summation over all data samples in Eqn. \ref{eqn_C3rdLadv}, \ref{eqn_C3rdLcls} and \ref{eqn_C3rdLreg} for simplicity.
Adversarial loss is to force the generated image be predicted as a real sample by $D_i$ and $C_i$. Meanwhile, we train the GAN discriminators $D$ to distinguish the evolving
fake images from the real ones and the loss function is:
\begin{equation}\small  \label{eqn_C3rdLcls}
\min_{D}L_{dis}=\sum_{i=1}^{n}\text{log}  D_i(X)+\text{log} (1- D_i(Y_{a}))
\end{equation}
A regularization loss is added to constrain the magnitude of noises:
\begin{equation}\small   \label{eqn_C3rdLreg}
\min_{E}L_{reg}=\left \| E(Y_{s})\right \|^{2}
\end{equation}

\subsection{Detection Track 1st}
\small{\emph{Members: Han Chen, Baoying Chen, Yanjie Hu, Shenghai Luo}}

\begin{figure}[htb]
\begin{center}
   \includegraphics[width=1.0\linewidth]{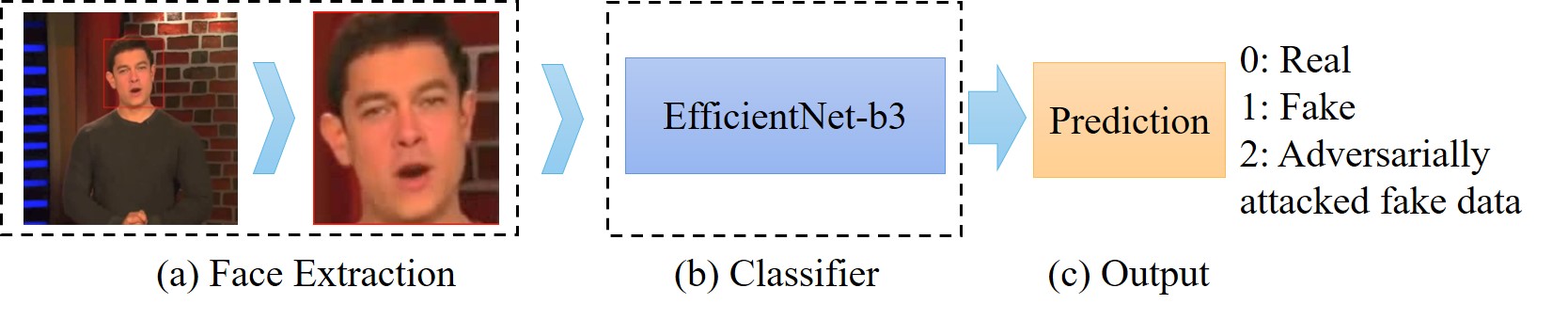}
\end{center}
   \caption{Detection track 1st solution's framework.}
\label{szu_fig1}
\end{figure}

Figure \ref{szu_fig1} illustrates the proposed overall framework. First, a face detector MTCNN \cite{zhang2016joint} is used to crop the face images from each video frame (enlarged the face region by a factor of 1.3). Then, EfficientNet-B3 \cite{tan2019efficientnet} as the backbone for forgery detection. Finally, the probability of real face is obtained. And this team's innovative solutions and training details are given as follows.

\begin{figure}[htb]
\begin{center}
   \includegraphics[width=0.8\linewidth]{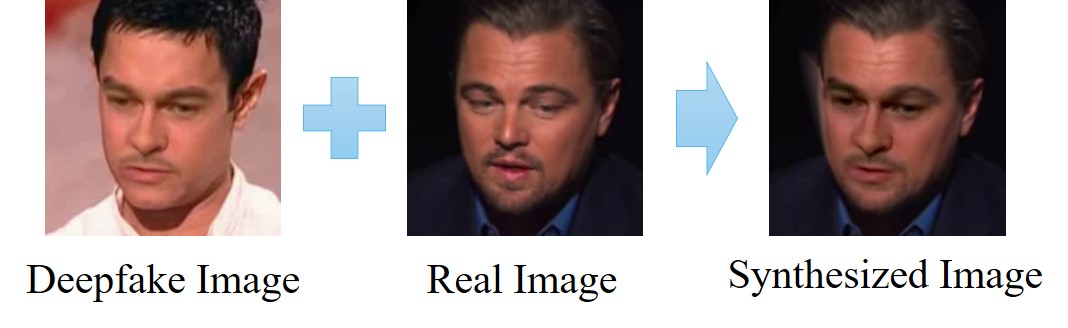}
\end{center}
   \caption{Detection track 1st solution's self-supervised forgery data generation.}
\label{szu_fig2}
\end{figure}

\textbf{Self-Supervised. }As the Face X-ray \cite{li2020face} said, in addition to detecting manipulation artifacts, blending artifacts can also be detected. So a similar method is employed to generate more forged face images. But the difference is that a Deepfake image is used as the foreground image, and the real image is used as the background image. The reasons are as follows: one is the face region of Deepfake image contains manipulation artifacts, which is very important for detection; and the other is that the number of Deepfake images in Celeb-DF \cite{CelebDF} is much larger than that of real images. Making full use of Deepfake images can synthesize more forged face images. Figure \ref{szu_fig2} illustrated the generated forged face images.
\\
\textbf{Adversarially attacked fake data. }Since the testing dataset contains a large number of adversarially attacked fake data, the team use some existing adversarial example algorithms to generate a part of adversarially attacked fake data for training. First of all, this team train lots of baseline model, e.g. VGG16 \cite{simonyan2014very} and ResNet18 \cite{he2016deep}.  Then the adversarial example algorithms employed include PGD \cite{madry2017towards} and MI-FGSM \cite{dong2018boosting}. In the end, due to the particularity of face images, we use two methods of to add noise, one is to add adversarial noise to the entire face image, and the other is to add adversarial noise only to the face region.
\\
\textbf{Multi-class Learning. }In order to make the network learn more distinguishing features, a multi-class learning method with three output neurons is employed. The $p_0\left ( x \right )$ denotes the probability of real face, $p_1\left ( x \right )$ denotes the probability of fake face and $p_2\left ( x \right )$ denotes the probability of adversarially attacked fake data. And the final probability of a fake face during the testing is calculated by $1-p_0\left ( x \right )$ or $\sum_{i=1}^{2}p_i\left ( x \right )$.
\\
\textbf{Implementation Details.}
This team employs a classifier based on EfficientNet-B3 with three class cross-entropy loss. The label smoothing technique is used to prevent the model from predicting the labels too confidently during training to improve generalization ability. A smoothing factor of 0.05 is used. Up-sampling technique is employed to balance the real and fake samples, and 75\% of the fake samples were adversarially attacked fake data. This team use SGD optimizer with initial learning rate of 0.001 and momentum of 0.9. The images are resized to $300 \times 300$, the batch size is set as 8, the total training epoch is 85, and the learning rate is reduced by 10\% every 2 epochs. The following data augmentations are applied during training: Gaussian noise, Gaussian blur, Horizontal flip and Self-supervised data augmentation mentioned above.

\subsection{Detection Track 2nd}
\small{\emph{Members: Junrui Huang, Yutong Yao, Boyuan Liu, Hefei Ling}}

\begin{figure}[h]
	\begin{center}
		\includegraphics[width=1\linewidth]{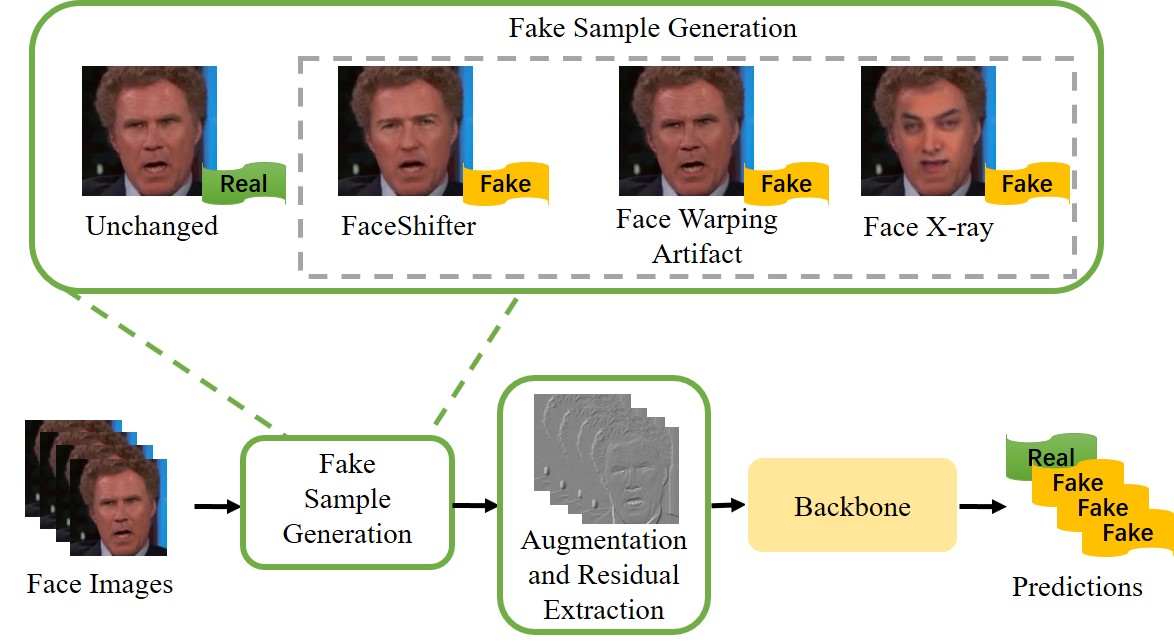}
		
	\end{center}
	\caption{Detection track 2nd solution's pipeline.}
	\label{fig:det-pipeline}
\end{figure}

As shown in Figure~\ref{fig:det-pipeline}, the detection method includes four stages: preprocessing, online fake sample generation, data augmentation, and a CNN based classifier.\\
{\bf Preprocessing.} we sample 50 frames evenly from each video. Faces are detected by MTCNN \cite{zhang2016joint}. We use a conservative crop (enlarged by a factor of 1.3) around the detected faces. Cropped faces are resized to $224\times224$ and saved as PNG format.\\
{\bf Online fake sample generation.} Since the training data is limited, we adopt three online fake sample generation methods to augment the training data, which are FaceShifter, Face Warping Artifacts, and Face X-ray. For the input real images, we use the three methods to generate fake samples, yet we keep the ratio of positive and fake samples to approximately 1:1.\\
$\bullet$ \emph{FaceShifter}: The team uses FaceShifter \cite{li2019faceshifter} pre-trained model for inference, for each input target image, we randomly pick a source image from the dataset to generate fake samples.\\
$\bullet$ \emph{Face Warping Artifacts}: Inspired by Face Warping Artifacts  \cite{li2018exposing}, the team simulates traditional face swapping, in which noise pattern differs between human face and background, by applying variable noises inside a general mask of face regions.\\ 
$\bullet$ \emph{Face X-ray}: The team refers to Face X-ray \cite{li2020face} and applies warp affine transformation from source face to target, followed by color transferring to achieve face-swapping.\\
{\bf Augmentation.} The team applies multiple traditional data augmentation methods including affine transformation, image compression, Gaussian blur, etc.. Besides, we use the solution from DFGC\footnote {https://github.com/selimsef/dfdc\_deepfake\_challenge} and randomly erase facial features from faces to improve generalization performance.\\
{\bf CNN-based classifier.} we use Efficientnet-b0 \cite{tan2019efficientnet} as the backbone model, and input images are resized to 224×224. Public pre-trained model\footnote {https://github.com/lukemelas/EfficientNet-PyTorch} is used instead of random init. Instead of direct RGB image input, we extract the edge of input faces to obtain high-frequency features.\\
{\bf Implementation Details.} In the training process, the batch size is set to 64. The learning rate is set to 0.0005 using Adam optimizer and then decayed 0.1 at 5000 iterations. Training stops at 6000 iterations to avoid over-fitting. For high-frequency noise extraction, we calculate the gray-scale image of the RGB input and apply convolution with edge detection kernel $[-1,1]$.
	
	
	
	

\subsection{Detection Track 3rd}
\small{\emph{Members: Changtao Miao, Changlei Lu, Shan He, Xiaoyan Wu, Wanyi Zhuang}}

\noindent \textbf{Dataset.}
For data preprocessing, we first extract all frames from each video using OpenCV. Then we apply the face detector MTCNN \cite{zhang2016joint} to detect the face region of each frame and expand the region by 1.3 times to crop the image.
In addition, we randomly sample the original real images in the Celeb-DF \cite{li2020celeb} dataset to use FaceShifier \cite{li2019faceshifter}, FSGAN \cite{nirkin2019fsgan} and First Order Motion \cite{siarohin2019first} to extend the corresponding fake images.
In general, the real samples are real images in the Celeb-DF \cite{li2020celeb}, and the fake samples include the fake images in Celeb-DF \cite{li2020celeb}, and the fake images generated by  FaceShifier \cite{li2019faceshifter}, FSGAN \cite{nirkin2019fsgan} and First Order Motion \cite{siarohin2019first}.
Meanwhile, we also use the FSGM \cite{goodfellow2014explaining} to generate adversarial samples of the part of training data, in order to increase the model's adversarial robustness. 



\noindent \textbf{Implementation Details.}
We employ EfficientNetV2 \cite{tan2021efficientnetv2} pre-trained on ImageNet as the backbone network.
And we apply the image-level data augmentation based on  Albumentations \cite{buslaev2020albumentations}, including: GaussianNoise, GaussianBlur, HueSaturationValue, IAAAdditiveGaussianNoise, IAASharpen, ISONoise, RandomBrightness, RandomBrightnessContrast.
Besides, we also try other data augmentation, such as: Augmix \cite{hendrycks2019augmix} and Cutmix \cite{yun2019cutmix}, however, the online test performance is not satisfactory, so we do not use it in the final submission model.
The images input size is 224$\times$224 or 288$\times$288, since we ensembled two models. The batch size is 128, and total training epoch is 4. 
In the training phase, we adopt the Adam optimizer with a learning rate of 0.0001 and weight decay of 0.001 and use a StepLR learning rate scheduler. And we balance the positive and negative
samples through the down-sampling technique.

\section{Further Analysis} \label{sec_Analysis}
We also conduct experiments to see whether high accuracy detection models on DFGC-test data generalize to other datasets, and whether high accuracy detection models trained on other datasets generalize to DFGC-test set.  Here DFGC-test are the submitted datasets from the final C-phase.

For the first analysis, we run the final D-phase models of DFGC on the test sets of DFDC \cite{DFDC} and FaceForensics++ \cite{FF++} datasets.
Results are shown in Table \ref{tabBeyond1}. The first column shows best detection AUC from final phase DFGC models. The second column shows the correlation coefficient between AUCs of DFGC models tested on DFGC-test and DFDC-test/FF++ test respectively. As can be seen, DFGC models still perform poorly on unseen datasets, and there is little correlation between detection performances on DFGC-test and those on unseen datasets.

For the second analysis, we run DFDC competition top-2 models on several different test sets, as shown in Table \ref{tabBeyond2}. DFDC-test and DFDC-wild are respectively the DFDC created test set that is similar to DFDC-train and the DFDC collected organic data. The DFDC top-2 models perform good on DFDC-test and Celeb-DF test sets, but are poor on in-the-wild data and even poorer on the DFGC-test data. Even though the DFDC models are trained on the largest DeepFake datasets to date, they are still vulnerable to unseen and adversarially attacked data.

We will release a majority part of the DFGC-test dataset as an extension of the Celeb-DF dataset at \url{https://github.com/yuezunli/celeb-deepfakeforensics}. We collected consents from 17 out of the 21 DeepFake datasets submitted to the final C-phase leaderboard. As discussed, these datasets include adversarial attacks, new fake methods and post-processings, and they pose a challenge to state-of-the-art detection models. This dataset can be used as a held-out testing set for evaluating the generalization ability and robustness of newly proposed detection methods.

\begin{table} [thb] 
\centering
\caption{DFGC models tested on two different datasets.}  \label{tabBeyond1}
\renewcommand{\arraystretch}{1.0}
\scalebox{0.8}{%
\begin{tabular}{|c|c|c|c|}
  \hline
       & Best DFGC Result   &Correlation \\
  \hline
DFDC-test   &0.682  &0.065  \\
\hline
FF++-test   &0.732 &0.189 \\
\hline
\end{tabular} 
}
\end{table}
\begin{table} [thb] 
\centering
\caption{DFDC top-2 models tested on different datasets. Metrics are AUROCs. }  \label{tabBeyond2}
\renewcommand{\arraystretch}{1.0}
\scalebox{0.8}{%
\begin{tabular}{|c|c|c|c|c|}
  \hline
    & DFDC-test  & DFDC-wild   & Celeb-DF   & DFGC-test \\
  \hline
DFDC-1st  &0.984  &0.717  & 0.904  &0.682 \\
\hline
DFDC-2nd  &0.985  &0.724 & 0.950  &0.696 \\
\hline
\end{tabular} 
}
\end{table}

The above results show that current DeepFake detection methods still struggle on generalizing to unseen datasets. Adversarial attacks also pose a serious problem to untargeted detection models. However, observing from the game procedure of this competition, relatively good detection performance can still be achieved when the counterpart information can be inferred or probed through multiple tries. This implies the importance of acquiring side-information in the DeepFake game, which can lead to more effective strategies than blindly designing more complicated models.

\section{Conclusion and Future Work}
In this paper, we introduce the organization of the DeepFake Game Competition, or DFGC, its overall design, protocols, evaluations and results. Top-3 solutions in both creation and detection tracks are also introduced. This competition is an initial attempt to evaluate DeepFake creation and detection in an adversarial and dynamic environment, which may mimic real-world challenges. 
We also release the DFGC-2021 dataset to the research community that can be used for testing new detection methods.

Throughout the three adversarial phases, the creation side improves in defeating the overall detection side, as discussed in Table \ref{tabStatistics}. This proves adversarial attacks and unseen DeepFake methods pose serious challenges to detection models. However, there are still some top detection solutions that can achieve quite high performance. The major trick that take effects may be the data augmentation in training with newly-created fake data and adversarially attacked data. This implies the importance of acquiring side-information and actively matching the training data distribution to that of the potential testing data. The cross-dataset analyses in Section \ref{sec_Analysis} shows low correlations between performances on the DFGC dataset and other datasets, which implies our competition dataset has very different distribution from existing datasets.

There are also some limitations in this competition  that we plan to improve in future work. First of all, we only let participants submit their created fake data, while the real data is not diverse enough. Secondly, submitted DeepFake datasets are also somewhat limited in diversity, concentrating on a small number of open-source DeepFake methods and heavily using adversarial attacks. Thirdly, training and testing data is not large. In future competitions, we may consider using or creating a much larger dataset and consider video inputs. Competition rules will be more carefully designed to encourage more diverse and high quality dataset submission. In summary, there are still many problems to be solved in the area of DeepFake detection and its evaluation, and we look forward to witness its progress together with the whole  community in future competitions.

\section*{Acknowledgements}
\noindent
We would like to thank \textit{Tianjin Academy for Intelligent Recognition Technologies} for sponsoring competition awards. \\
We would like to thank \textit{Alibaba Security} for sponsoring the evaluation server. \\
We would also like to thank the following participants for sharing their created DeepFake datasets to the research community: \\
\textit{Zhiliang Xu, Quanwei Yang, Fengyuan Liu, Hang Cai, Shan He, Christian Rathgeb, Daniel Fischer, Binghao Zhao, Li Dongze.}

{\footnotesize
\bibliographystyle{ieee}
\bibliography{egbib}
}

\end{document}